\documentclass[10pt,twocolumn,letterpaper]{article}

\usepackage[pagenumbers]{cvpr} % To force page numbers, e.g. for an arXiv version

\definecolor{cvprblue}{rgb}{0.21,0.49,0.74}
\usepackage[pagebackref,breaklinks,colorlinks,allcolors=cvprblue]{hyperref}
\usepackage{multirow}
\usepackage[accsupp]{axessibility}
\usepackage{amssymb}
\usepackage{mathrsfs}
\usepackage{marvosym}
\def\paperID{30864} % *** Enter the Paper ID here
\def\confName{CVPR}
\def\confYear{2026}

\title{DynamicVGGT: Learning Dynamic Point Maps for 4D Scene Reconstruction in Autonomous Driving}

\author{
Zhuolin He$^{1,3}$ \quad
Jing Li$^{2}$ \quad
Guanghao Li$^{1,4}$ \quad
Xiaolei Chen$^{1}$ \quad
Jiacheng Tang$^{1}$ \quad
Siyang Zhang$^{3,5}$ \\
Zhounan Jin$^{2}$ \quad
Feipeng Cai$^{3}$ \quad
Bin Li$^{1}$ \quad
Jian Pu$^{1\text{\Letter}}$ \quad
Jia Cai$^{3}$ \quad
Xiangyang Xue$^{1\text{\Letter}}$ \\
$^{1}$Fudan University \quad
$^{2}$Huawei \quad
$^{3}$Yinwang Intelligent Technology \\
$^{4}$Shanghai Innovation Institute \quad
$^{5}$CUHK \\
\tt\small{ \{zlhe22,ghli22,chenxl23,jiachengtang21\}@m.fudan.edu.cn} \\ \tt\small{\{libin,jianpu,xyxue\}@fudan.edu.cn \{lijing470,jinzhounan\}@huawei.com} \\
 \tt\small{\{caifeipeng,caijia\}@yinwang.com siyangzhang@link.cuhk.edu.cn}
}
\begin{document}

\maketitle

\begin{abstract}
Dynamic scene reconstruction in autonomous driving remains a fundamental challenge due to significant temporal variations, moving objects, and complex scene dynamics. 
Existing feed-forward 3D models have demonstrated strong performance in static reconstruction but still struggle to capture dynamic motion. 
To address these limitations, we propose DynamicVGGT, a unified feed-forward framework that extends VGGT from static 3D perception to dynamic 4D reconstruction. 
Our goal is to model point motion within feed-forward 3D models in a dynamic and temporally coherent manner. 
To this end, we jointly predict the current and future point maps within a shared reference coordinate system, allowing the model to implicitly learn dynamic point representations through temporal correspondence. 
To efficiently capture temporal dependencies, we introduce a Motion-aware Temporal Attention (MTA) module that learns motion continuity. 
Furthermore, we design a Dynamic 3D Gaussian Splatting Head that explicitly models point motion by predicting Gaussian velocities using learnable motion tokens under scene flow supervision. It refines dynamic geometry through continuous 3D Gaussian optimization. 
Extensive experiments on autonomous driving datasets demonstrate that DynamicVGGT significantly outperforms existing methods in reconstruction accuracy, achieving robust feed-forward 4D dynamic scene reconstruction under complex driving scenarios.

% We will release the code and pretrained models upon acceptance.
\end{abstract}

% hzl - 1022edit
% v2 从 参考 page-4d改
% 在自动驾驶中，动态场景重建仍然是一项基础且具有挑战性的任务，这主要源于时间维度上的剧烈变化、移动物体的存在以及动态场景对静态几何假设的违背。现有的前馈式三维模型（如 Visual Geometry Grounded Transformer，VGGT）在静态场景重建中表现优异，但在处理时序一致性和动态运动方面存在不足。为此，我们提出了 DynamicVGGT，一个将 VGGT 从静态三维感知扩展至动态四维重建的统一前馈框架。DynamicVGGT 的核心在于一种 动态点图（Dynamic Point Map）表示，该机制在共享参考坐标系下联合预测当前帧与未来帧的三维点图，其位移可用于粗略估计场景流并建模动态几何变化。为了高效捕获时序依赖关系，我们进一步引入了 并行时间注意力（Parallel Temporal Attention） 模块，以学习跨帧的长程运动连续性；同时设计了基于高斯的 场景变形与点追踪（GSDPT） 模块，通过连续三维高斯优化对动态几何进行精细重建。大量在合成与真实自动驾驶数据集上的实验表明，DynamicVGGT 在重建精度与时序一致性方面均取得显著提升，能够在复杂驾驶场景中实现鲁棒的前馈式四维动态场景重建。
\section{Introduction}
\label{sec:intro}

% 背景和现实需求
Visual geometry learning is a fundamental problem in computer vision and serves as a core foundation for various applications in robotics ~\cite{yan2025ordermind} and autonomous driving ~\cite{WAM-Diff-2025, WAM-Flow-2025}. In recent years, feed-forward 3D models~\cite{driv3r, smart2024splatt3r, tian2025drivingforward} have achieved remarkable progress in static scene understanding by directly predicting geometric representations such as point clouds and 3D Gaussian from image inputs.  However, visual geometry learning in autonomous driving scenarios faces much greater complexity than in static scenes. Real-world driving environments are inherently dynamic, featuring diverse moving objects, changing long-range temporal dependencies. Although feed-forward architectures~\cite{wang2025vggt, wang2025moge} demonstrate strong performance on static datasets, they struggle to maintain both geometric accuracy and temporal consistency when extended to such dynamic conditions.
This motivates the need for a unified feed-forward framework that can jointly model geometry and motion, enabling temporally consistent dynamic scene reconstruction.

\begin{figure}[t]
\centering
    \includegraphics[width=0.95\columnwidth]{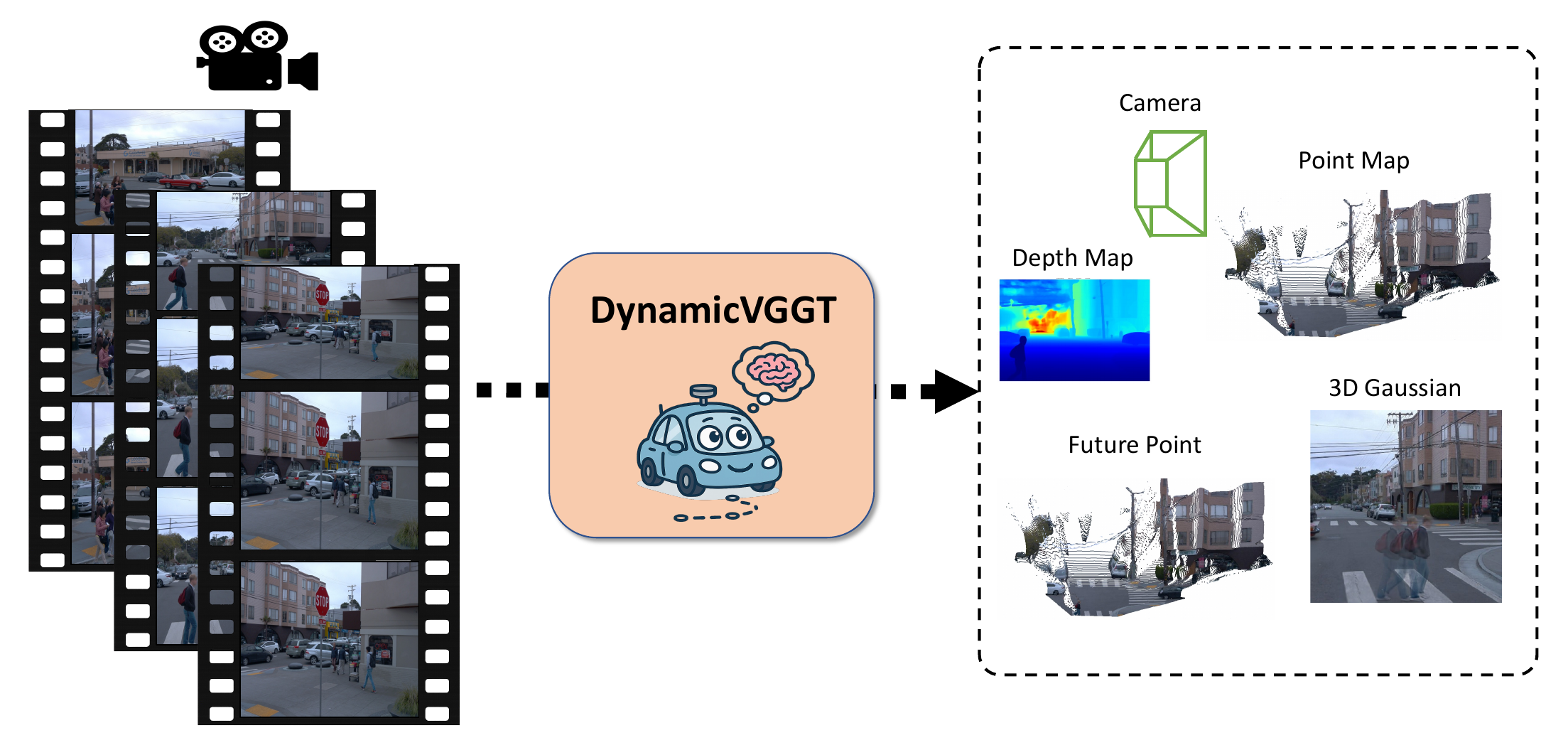}
    \vspace{-2mm}
    \captionof{figure}{
        \textbf{DynamicVGGT} extends static multi-view 3D perception to dynamic 4D reconstruction by enabling 3D Gaussian rendering and adaptively modeling motion across multiple temporal scales without explicit camera extrinsic alignment.
    }
    \label{fig:framework}
\end{figure}

Current 3D foundation models~\cite{fan2024instantsplat, jiang2025anysplat} are typically trained on large-scale, well-labeled datasets and can achieve consistent and accurate 3D reconstruction across most scenes.
However, applying them to real-world autonomous driving scenarios remains highly challenging.
First, autonomous driving data often exhibit large-scale, high-noise, and sparse-depth characteristics, which can lead to a degradation of the model’s original dense prediction capability when trained directly on such data.
Moreover, beyond static geometric perception, capturing dynamic geometric information is crucial in autonomous driving.
Although several recent 3D foundation models~\cite{zhuo2025streaming, lin2025movies} have begun to explore dynamic scene modeling, their output representations are still primarily based on static point maps, lacking a unified dynamic representation that can directly support downstream autonomous driving tasks.

To address these issues, we propose DynamicVGGT, a unified framework for high-fidelity dynamic scene reconstruction in a feed-forward manner. 
As Fig.~\ref{fig:framework} shows, DynamicVGGT introduces a novel Dynamic Point Map (DPM) \cite{sucar2025dynamic} mechanism designed by two different dynamic tasks.
Specifically, we introduce a Future Point Head that predicts the point map of the next frame and enforces consistency with the current frame, thereby enabling the model to implicitly learn point-wise motion. On the other hand, we introduce a Dynamic 3D Gaussian Splatting Head (DGSHead), which refines the predicted geometry using Gaussian primitives initialized from the geometric priors. 
It further incorporates a lightweight motion-aware encoder that encodes motion flow through learnable motion tokens, supervised by scene flow.
Extensive experiments demonstrate that DynamicVGGT achieves state-of-the-art performance across diverse driving datasets.  We summarize our main contributions as follows:
\begin{itemize}

\item We introduce a motion-aware temporal attention module that captures temporal dependencies without disrupting VGGT’s spatial attention, preserving stable training and geometric priors.

\item We extend point-based representations towards a unified DPM by introducing a future point prediction task and a Dynamic 3D Gaussian Splatting Head.
On top of this framework, the model learns point-wise motion through implicit consistency of inter-frame point motion and explicit supervision of Gaussian motion using scene flow.

\item A stage-wise training scheme is adopted to mitigate the performance degradation observed on real-world driving data.
On the Waymo dataset, our model achieves notable gains over VGGT and StreamVGGT, improving Accuracy by 0.5 and Completeness by 0.2.
\end{itemize}

\section{Related work}
\label{sec:related_work}
\begin{figure*}[t]
\centering
  \includegraphics[width=0.95\textwidth]{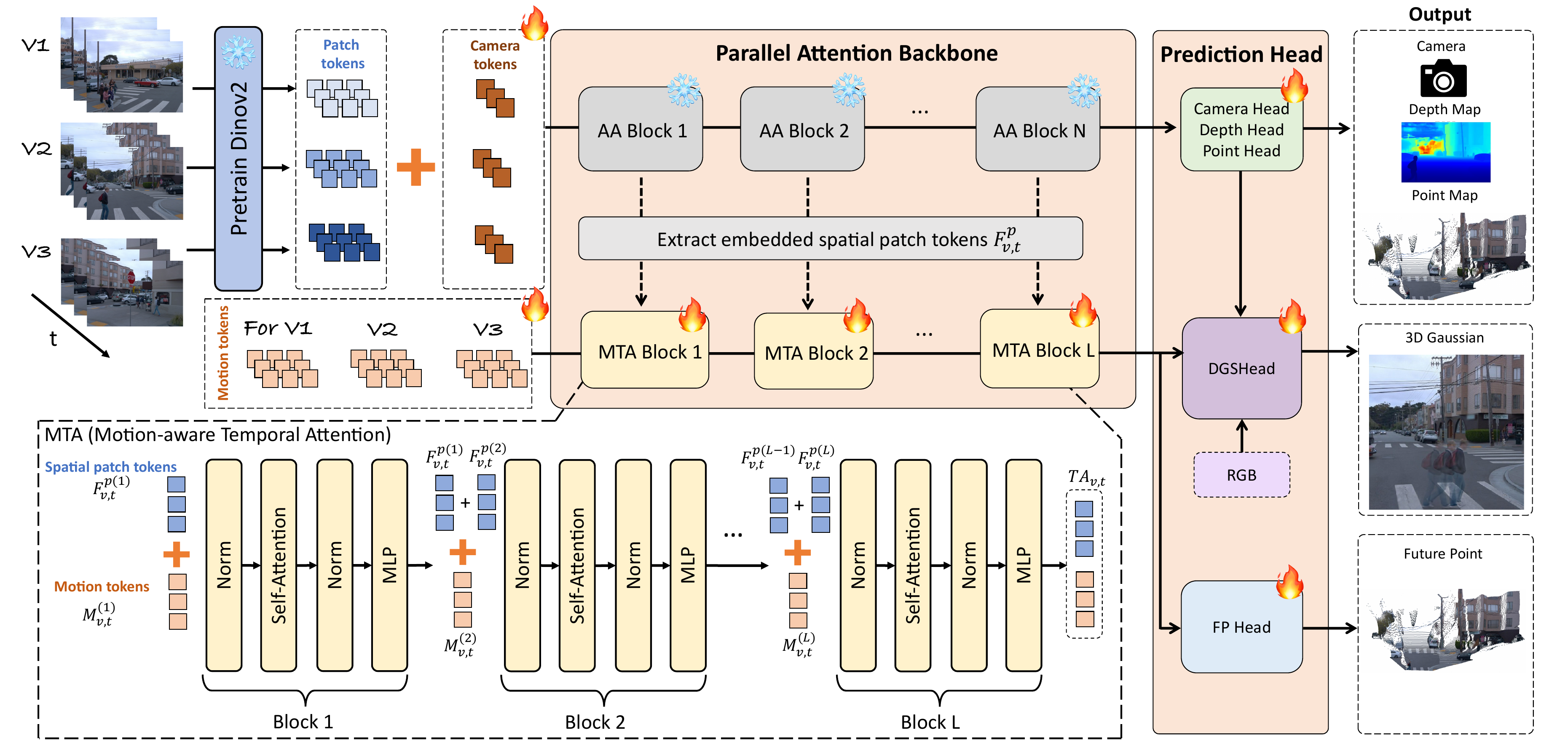}
  \caption{ \textbf{Proposed DynamicVGGT training framework.} Given a sequence of multi-view images $\{V_1,V_2,V_3\}$, the model first encodes them using a pretrained DINOv2 backbone to extract patch tokens and camera tokens for each view, while motion tokens are initialized as learnable parameters that encode temporal priors. 
  The patch and camera tokens are processed by the Alternating-Attention (AA) blocks to model intra-frame spatial geometry, 
  whereas the Motion-aware Temporal Attention (MTA) blocks operate in parallel to model inter-frame temporal dependencies using the motion tokens.
  The resulting temporal features ${TA}$ are then fed into a Dynamic 3D Gaussian Head (DGSHead) for dynamic 3DGS reconstruction and a Future Point Head for future point prediction.
  }
  \label{fig:pipeline}
\end{figure*}
\subsection{Feed-Forward Visual Geometry Learning.}
% [介绍 DUSt3R, VGGT 等前馈几何模型的发展与局限]
Feed-forward 3D reconstruction models~\cite{wang2024dust3r, wang2025vggt, jiang2025anysplat} aim to recover the geometry of static scenes directly from images under a temporal invariance assumption, providing robust 3D priors for downstream tasks~\citep{chen2024vpl, li2025papl, li2025constrained, liartdeco, li2026ec, gaodeep, gaogood}.
Unlike traditional multi-view geometry pipelines~\cite{mur2015orb, mur2017orb} that relied on optimization-based correspondence matching, these learning-based methods predicted depth, camera pose, or dense 3D point maps in an end-to-end manner. 
Representative frameworks such as DUSt3R~\cite{wang2024dust3r} had demonstrated that transformer-based architectures can effectively learn direct mappings from image pixels to 3D coordinate fields. Subsequent researches had expanded this paradigm to multi-view and sequential settings, integrating camera pose estimation, depth prediction, and correspondence learning into unified architectures. 
Among these, VGGT~\cite{wang2025vggt} further enhanced the feed-forward formulation by introducing alternating attention mechanisms across spatial and temporal dimensions, achieving joint prediction of multiple geometric quantities through a single, shared model. 
Anysplat~\cite{jiang2025anysplat} incorporated 3DGS with Feed-Forward Model to achieve high-fidelity reconstruction. Recent extensions of feed-forward visual geometry models have begun to incorporate temporal modeling for dynamic reconstruction. Approaches such as MoVieS\cite{lin2025movies} and StreamVGGT\cite{zhuo2025streaming} extend static frameworks like VGGT to handle sequential inputs, but they are primarily designed for indoor environments.
Despite these advances, existing 3D feed-forward methods struggled with reconstruct the large scale, dynamic autonomous driving scenes, motivating the need for generalizable 4D feed-forward frameworks capable of capturing scene dynamics over time.

\subsection{3DGS Reconstruction for Driving Scenes.}
% [介绍动态 3DGS 方法在自动驾驶上的相关工作， DynamicVGGT做自动驾驶数据集]
Building photorealistic reconstructions of dynamic urban scenes is crucial for autonomous driving, as it supports closed-loop training and evaluation under realistic motion patterns. Consequently, recent research~\cite{yan2024street} has shifted focus from static to dynamic scene reconstruction. However, most existing methods~\cite{chen2024omnire, zhou2024hugs} rely heavily on dense annotations, which are expensive to obtain and limit scalability. Furthermore, these approaches typically depend on per-scene optimization, making it difficult to exploit large-scale data priors and leading to slow reconstruction speeds.
Recent advances in feed-forward reconstruction, such as STORM~\cite{yang2024storm} and DrivingForward~\cite{tian2025drivingforward}, demonstrate the potential for fast and generalizable 4D scene recovery without per-scene optimization.
STORM introduces a feed-forward pipeline for dynamic scene reconstruction and editing, achieving high-quality results but still relying on calibrated multi-view inputs.
DrivingForward~\cite{tian2025drivingforward} presents a feed-forward 3D Gaussian Splatting framework for driving scenes with flexible surround-view inputs, jointly training pose, depth, and Gaussian heads to infer camera poses and dense geometry without using depth ground truth or provided extrinsics. Our DynamicVGGT generalizes feed-forward reconstruction to real-world autonomous driving scenes, jointly modeling geometry and motion without camera parameters or dense annotations.

\section{Method}

Our framework builds upon VGGT and extends it from static 3D perception to dynamic 4D reconstruction. 
The key idea is to establish a unified geometric representation, namely DPMs, as the core of temporal modeling. 
Based on this formulation, we introduce temporal reasoning via a MTA module, predict future geometry via a Future Point Head (FPH), and further refine dynamic geometry through a DGSHead. 
An overview of the proposed architecture is shown in Fig.~\ref{fig:pipeline}.

\subsection{Dynamic Point Map and Task Formulation}

We denote the camera index by $v \in \{1,\dots,N_v\}$ and the frame index by $t \in \{1,\dots,\tau\}$. 
Temporal modeling is defined on frame pairs $(v,t)$ and $(v,t+\delta)$ from the same camera stream.
Following prior work~\cite{sucar2025dynamic}, a static point map is defined as
\begin{equation}
P_{v,t} = \pi^{-1}(I_{v,t}; K_{v,t}, E_{v,t}) \in \mathbb{R}^{3 \times H \times W}.
\end{equation}

To model dynamics, previous DPM formulations align all frames into a shared reference frame:
\begin{equation}
P_{v,t}^{(\mathrm{ref})} =
\mathcal{T}_{(v,t)\rightarrow \mathrm{ref}}
\big(\pi^{-1}(I_{v,t};K_{v,t},E_{v,t})\big),
\end{equation}

so that temporal motion is expressed as
\begin{equation}
\Delta P_{v,t}^{(\mathrm{ref})} =
P_{v,t+\delta}^{(\mathrm{ref})}-P_{v,t}^{(\mathrm{ref})}.
\end{equation}

In contrast, VGGT directly predicts point maps in a learned canonical frame. 
Given a multi-view clip $\{I_{v,t}\}$, our model predicts
\begin{equation}
\hat{P}_{v,t},\hat{P}_{v,t+\delta}=
f_\theta(\{I_{v,t}\})|_{(v,t),(v,t+\delta)},
\end{equation}

which enables implicit motion learning through
$
\Delta\hat{P}_{v,t}=\hat{P}_{v,t+\delta}-\hat{P}_{v,t}.
$
This formulation avoids explicitly relying on externally specified frame-to-reference transformations in the dynamic formulation, while still preserving the geometric prior of the original VGGT backbone. 

\begin{figure}[t]
\centering
  \includegraphics[width=0.9\columnwidth]{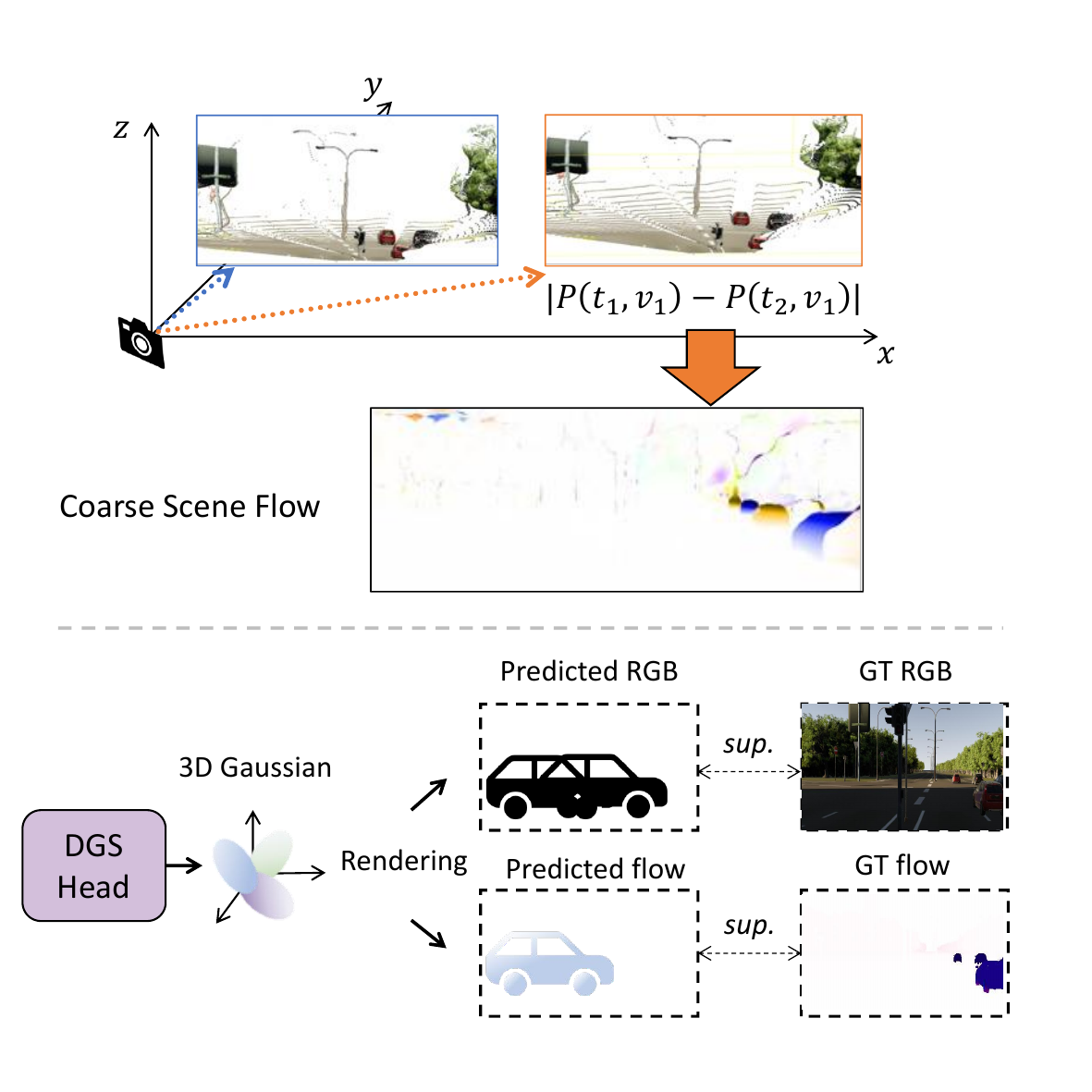}
  \caption{\textbf{Dynamic task formulation.} 
  We formulate dynamic point maps by designing two complementary tasks that model point-wise motion over time. 
  The Future Point Head learns implicit motion through inter-frame point consistency, while the Dynamic 3D Gaussian Splatting Head provides explicit motion supervision via scene flow to refine dynamic geometry.}
  \label{fig:dynamic_task}
\end{figure}

It therefore provides a unified basis for dynamic task design, as illustrated in Fig.~\ref{fig:dynamic_task}.
Specifically, we introduce two complementary tasks built upon this formulation: 
(i) the FPH, which learns implicit motion by enforcing inter-frame point consistency, and 
(ii) the DGSHead, which explicitly refines dynamic geometry through scene-flow-supervised Gaussian optimization.

\subsection{Motion-aware Temporal Attention}

Although the DPM formulation captures temporal variations at the geometric level, relying solely on point-wise displacement is still insufficient for accurate dynamic reconstruction. 
Previous work, such as StreamVGGT~\cite{zhuo2025streaming}, also introduces temporal attention to enhance temporal modeling. 
However, its sequential stacking of AA blocks tends to cause unstable training and degraded performance in the early stages.
To address this limitation, we propose a Motion-aware Temporal Attention (MTA) module that explicitly models motion cues at the feature level and integrates temporal reasoning into the VGGT backbone. 
The key idea is to introduce learnable motion tokens that dynamically encode inter-frame motion information, guiding temporal attention to focus on motion-consistent regions.

Given the aggregated token features $\tilde{F}_{v,t} = [F_{v,t}^{c}; F_{v,t}^{p}]$ produced by the AA blocks, we remove the camera token $F_{v,t}^{c}$ and concatenate the patch tokens $F_{v,t}^{p}$ with learnable motion tokens to form the MTA input. 
For each MTA layer $l$, temporal correlations are computed in parallel across all frames along the temporal dimension $\tau$. 
The input to the $l$-th MTA layer is defined as
\begin{equation}
F_{m,v,t}^{(l)} = 
\begin{cases}
\mathrm{Concat}\!\left(M_{v,t}^{(l)},\, F_{v,t}^{p(l)}\right), & l = 1,\\
\mathrm{Concat}\!\left(M_{v,t}^{(l)},\, F_{v,t}^{p(l)} + F_{v,t}^{p(l-1)}\right), & l > 1,
\end{cases}
\end{equation}
where $M_{v,t}^{(l)}$ denotes the motion tokens and $F_{v,t}^{p(l)}$ denotes the spatial patch tokens from the AA branch at layer $l$.
% 中文：这里把所有 token 都补上了 view index。
% 中文：同时把时间长度 T 改成 \tau，避免和 transformation T 混淆。
Temporal attention is computed independently for each patch position and each view:
\begin{equation}
A_{t,t'}^{(l)} = 
\mathrm{Softmax}\!\left(
\frac{Q_{t}^{\mathrm{attn},(l)} \left(K_{t'}^{\mathrm{attn},(l)}\right)^\top}{\sqrt{d}}
+ B_{t,t'}^{\mathrm{time}}
\right),
\end{equation}
\begin{equation}
\tilde{F}_{m,v,t}^{(l)} = \sum_{t'=1}^{\tau} A_{t,t'}^{(l)} V_{t'}^{\mathrm{attn},(l)},
\end{equation}
where $t,t' \in \{1,\dots,\tau\}$ denote discrete frame indices within the sampled clip, and $B_{t,t'}^{\mathrm{time}}$ denotes the temporal positional bias implemented using rotary position embeddings.
The updated temporal features are then processed by layer normalization and an MLP with residual connections:
\begin{equation}
F_{m,v,t}^{(l+1)} = \mathrm{MLP}^{(l)}\!\left(\mathrm{LayerNorm}\!\left(\tilde{F}_{m,v,t}^{(l)}\right)\right) + F_{m,v,t}^{(l)}.
\end{equation}

After the final MTA layer, we denote the temporally enhanced feature for view $v$ at time $t$ by
\begin{equation}
TA_{v,t} = F_{m,v,t}^{(L)},
\end{equation}
where $L$ is the number of MTA layers. 
The resulting feature $TA_{v,t}$ is used by both the Future Point Head and the Dynamic 3DGS Head.
This formulation enables simultaneous message passing across temporal spans and enhances the model’s ability to capture motion continuity and temporally coherent geometry.

\subsection{Future Point Prediction}

Building upon the unified DPM representation and the feature-level temporal modeling provided by MTA, 
we further introduce a Future Point Head (FPH) to explicitly learn point-wise motion dynamics. 
Specifically, the FPH predicts the 3D point map of a future frame from the temporally enhanced feature at the current timestep, enabling the network to learn short-term motion continuity in a self-supervised manner.
Given the temporally enhanced feature $TA_{v,t}$ produced by the MTA module, the FPH predicts the point map of the same camera stream at a future timestep:
\begin{equation}
\hat{P}_{v,t+\delta}^{\,\mathrm{fut}} = \mathrm{DPT}_{p}(TA_{v,t}),
\end{equation}
where $\mathrm{DPT}_{p}(\cdot)$ denotes a DPT head~\cite{ranftl2021vision} for future point regression.
To further encourage physically plausible point motion trajectories, we introduce a temporal consistency regularization:
\begin{equation}
\mathcal{L}_{\mathrm{temp}}=
\frac{1}{|\mathcal{N}|}\sum_{i\in \mathcal{N}}
\left\|
\left(\mathbf{p}_{v,t+\delta}^{(i)}-\mathbf{p}_{v,t}^{(i)}\right)
-
\left(\hat{\mathbf{p}}_{v,t+\delta}^{(i)}-\hat{\mathbf{p}}_{v,t}^{(i)}\right)
\right\|_{1},
\label{eq:temporal_consistency}
\end{equation}
where $\mathbf{p}_{v,t}^{(i)}$ and $\hat{\mathbf{p}}_{v,t}^{(i)}$ denote the ground-truth and predicted 3D coordinates of the $i$-th valid point at time $t$, respectively, and $\mathcal{N}$ denotes the set of valid points.
The displacement field
$
\Delta \mathbf{p}_{v,t}^{(i)} = \mathbf{p}_{v,t+\delta}^{(i)}-\mathbf{p}_{v,t}^{(i)}
$
acts as a coarse motion representation that encourages the network to learn inter-frame point displacement within the shared DPM coordinate space. 
We emphasize that $\mathcal{L}_{\mathrm{temp}}$ supervises motion implicitly at the point-map level, which complements the explicit motion supervision introduced later in the Dynamic 3DGS Head.

\subsection{Dynamic 3D Gaussian Splatting Head}

To further model dynamic scenes at the primitive level, we introduce a Dynamic 3D Gaussian Splatting (3DGS) Head as a downstream dynamic reconstruction task.

This module takes both the temporally enhanced features $TA_{v,t}$ from MTA and the RGB appearance cues from the input images as inputs, and converts them into time-varying 3D Gaussian primitives that jointly model geometry, appearance, and motion. 
We observe that freezing the AA blocks causes the pretrained VGGT backbone to overemphasize geometric reasoning while weakening appearance cues, which degrades Gaussian rendering quality. 
To compensate for this issue, we fuse image features extracted by $\mathrm{Conv}(\cdot)$ with geometric features obtained from $TA_{v,t}$:
\begin{equation}
F_{v,t}^{\mathrm{app}} = \mathrm{Conv}(I_{v,t}),
\end{equation}
\begin{equation}
F_{g,v,t},\, D_{g,v,t} = \mathrm{DPT}_{g}(TA_{v,t}),
\end{equation}
\begin{equation}
G_{v,t} = F_{v,t}^{\mathrm{app}} + F_{g,v,t},
\end{equation}
where $F_{g,v,t}$ denotes the Gaussian features used to initialize Gaussian primitives, and $D_{g,v,t}$ denotes the predicted Gaussian depth.
At each timestep, the predicted Gaussian depth $D_{g,v,t}$ together with the camera predictions from the retained VGGT camera branch are used to reconstruct a point map $P_{v,t}^{g}$, which initializes the Gaussian centers $\mu_i$. 
Each Gaussian primitive is parameterized as
$
\{\mu_i, \sigma_i, r_i, c_i, \nu_i\},
$
where $\mu_i$ denotes the center, $\sigma_i$ the scale, $r_i$ the rotation, $c_i$ the color, and $\nu_i$ the velocity vector.

We further leverage the $M$ learnable motion tokens introduced in MTA to decode a set of velocity bases $\nu_b \in \mathbb{R}^{3}$, forming a shared dynamic representation for Gaussian motion. 
To describe temporal evolution, we assume constant velocity within a short clip:
\begin{equation}
\mu_{i,t+\delta} = \mu_{i,t} + \delta \cdot \nu_{i,t}.
\end{equation}

During training, scene-flow supervision is applied to encourage each Gaussian primitive to carry a physically meaningful velocity attribute. 
Unlike $\mathcal{L}_{\mathrm{temp}}$, which constrains coarse inter-frame displacement at the point-map level, the Gaussian motion supervision explicitly regularizes motion in the dynamic Gaussian space and therefore provides complementary supervision for dynamic reconstruction.

% 中文：这句再次明确 L_temp 和 L_flow 分工不同，不是重复监督。

\subsection{Training Objective}

To enable the model to learn dynamic geometry progressively, we adopt a two-stage training strategy that follows a curriculum-style paradigm from synthetic to real-world data. 
In the first stage, the model is trained on high-quality synthetic autonomous driving datasets, where dense geometry and reliable motion cues are available. 
The stage-1 training objective combines static and temporal supervision:
\begin{equation}
\mathcal{L}_{\mathrm{stage1}} =
\mathcal{L}_{\mathrm{cam}}
+ \mathcal{L}_{\mathrm{depth}}
+ \mathcal{L}_{\mathrm{point}}^{(t)}
+ \mathcal{L}_{\mathrm{point}}^{(t+\delta)}
+ \lambda_{\mathrm{temp}} \mathcal{L}_{\mathrm{temp}}.
\end{equation}

The camera loss supervises the predicted camera parameters $\hat{g}$:
$
\mathcal{L}_{\mathrm{cam}} = \sum_{i=1}^{N} \left\| \hat{g}_i - g_i \right\|_{\epsilon},
$
where $\hat{g}_i$ and $g_i$ denote the predicted and ground-truth camera parameters, respectively, and $\|\cdot\|_{\epsilon}$ denotes the Huber loss. 
The depth loss $\mathcal{L}_{\mathrm{depth}}$ and point-map losses $\mathcal{L}_{\mathrm{point}}^{(t)}$ and $\mathcal{L}_{\mathrm{point}}^{(t+\delta)}$ follow VGGT~\cite{wang2025vggt}. 
This stage focuses on learning temporal consistency by predicting future point maps, allowing the network to capture short-term motion while preserving the geometric priors of the pretrained backbone.

In the second stage, we fine-tune the model on real driving datasets using the Dynamic 3DGS objective. 
The overall stage-2 objective is defined as
\begin{equation}
\mathcal{L}_{\mathrm{stage2}} = \mathcal{L}_{\mathrm{stage1}} + \mathcal{L}_{\mathrm{3DGS}},
\end{equation}
where
\begin{equation}
\mathcal{L}_{\mathrm{3DGS}} =
\mathcal{L}_{\mathrm{rgb}}
+ \lambda_{\mathrm{gs}} \mathcal{L}_{\mathrm{gsdepth}}
+ \lambda_{\mathrm{dist}} \mathcal{L}_{\mathrm{distill}}
+ \lambda_{\mathrm{flow}} \mathcal{L}_{\mathrm{flow}}.
\end{equation}

\begin{figure}[t]
\centering
  \includegraphics[width=0.9\columnwidth]{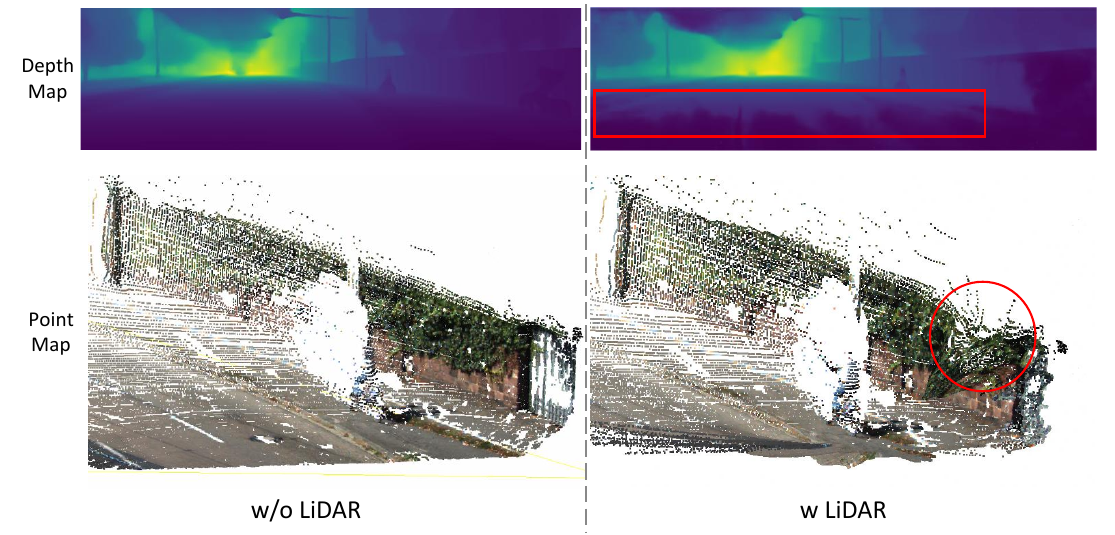}
  \caption{\textbf{Depth and Point Maps Comparison}. The sparsity of LiDAR point clouds degrades the results, leading to less smooth depth maps and rougher point clouds.}
  \label{fig:depth_lidar}
\end{figure}

For image reconstruction, we use
$
\mathcal{L}_{\mathrm{rgb}} = \mathrm{MSE}(I_{v,t}, \hat{I}_{v,t}),
$
where $I_{v,t}$ is the ground-truth RGB image and $\hat{I}_{v,t}$ is the image rendered from the predicted dynamic Gaussian primitives.

We observe that directly using sparse LiDAR point clouds as supervision on real autonomous driving datasets leads to severe performance degradation due to their limited density and uneven spatial distribution. 
To alleviate this issue, we introduce a depth distillation strategy. 
Specifically, the depth predicted by the stage-1 point-map branch serves as the teacher signal, while the Gaussian depth branch predicts the student depth:
$
\mathcal{L}_{\mathrm{distill}} = \left\| D_{g,v,t} - \mathrm{sg}\!\left(D^{\mathrm{pm}}_{v,t}\right) \right\|_1,
$
where $D^{\mathrm{pm}}_{v,t}$ denotes the depth predicted by the stage-1 geometry branch and $\mathrm{sg}(\cdot)$ denotes the stop-gradient operator. 
In addition, $\mathcal{L}_{\mathrm{gsdepth}}$ is a standard $L_1$ loss, whose supervision is provided by the pretrained model~\cite{wang2025moge}. 
This strategy mitigates the noise caused by point-cloud sparsity and stabilizes Gaussian optimization, as shown in Fig.~\ref{fig:depth_lidar}.

\begin{table*}[t]
\centering
\caption[Valori medi]{\textbf{Point Map Reconstruction on KITTI and Waymo(val).} KITTI uses monocular input with every 3 consecutive frames per camera. Waymo uses 3 frames (stride 4) from FRONT, SIDE\_LEFT, and SIDE\_RIGHT cameras, totaling 9 images per group. }
{\footnotesize\setlength{\tabcolsep}{.8mm}
{\begin{tabular}{lcccccccccccccccccc}
\toprule[1.2pt]                      
\multirow{3}{*}{\textbf{Methods}} & \multicolumn{6}{c}{\textbf{{KITTI (monocular)}}}   & \multicolumn{6}{c}{\textbf{{Waymo (3 cameras)}}}
\\ 
\cmidrule(lr){2-7} \cmidrule(lr){8-13} \cmidrule(lr){14-19}
& \multicolumn{3}{c}{\textbf{{Mean}}} & \multicolumn{3}{c}{\textbf{{Med.}}} &
\multicolumn{3}{c}{\textbf{{Mean}}} & \multicolumn{3}{c}{\textbf{{Med.}}} 
\\ 
\cmidrule(lr){2-4} \cmidrule(lr){5-7} \cmidrule(lr){8-10} \cmidrule(lr){11-13}
          & $ \text{Acc.} \downarrow $     & 
          $ \text{Comp.} \downarrow  $     &
          $ \text{NC} \uparrow  $     & 
          $ \text{Acc.} \downarrow   $     & 
          $ \text{Comp.} \downarrow    $     & 
          $ \text{NC} \uparrow   $     & 
          $ \text{Acc.} \downarrow   $     & 
          $ \text{Comp.} \downarrow    $     &
          $ \text{NC} \uparrow  $     & 
          $ \text{Acc.} \downarrow $     & 
          $ \text{Comp.} \downarrow  $     &
          $ \text{NC} \uparrow  $     &   
\\ 
\midrule

VGGT\cite{wang2025vggt}            & 1.489 & 0.690 & 0.918 & 1.329 & \textbf{0.535} & \textbf{0.971}    & 4.635 & 2.667 & 0.561 & 2.634 & 1.734 & 0.590  \\
StreamVGGT\cite{zhuo2025streaming}      & 1.078 & \textbf{0.495} & 0.899 & 0.867 & 0.390 & 0.949    & 4.598 & 2.626 & 0.564 & 2.567 & 1.789 & 0.592  \\
DynamicVGGT     &\textbf{0.901} &\underline{0.584} &\textbf{0.939} &\textbf{0.733} &\underline{0.464} &\underline{0.963}   &\textbf{4.021} &\textbf{2.390} &\textbf{0.562} &\textbf{1.971} &\underline{1.564} &\underline{0.603}                \\

\bottomrule[1.2pt]
\end{tabular}
}
}
\label{tab:point-estimation}
\end{table*}

Finally, we use scene-flow supervision for explicit Gaussian motion learning:
$
\mathcal{L}_{\mathrm{flow}} = \mathrm{MSE}(\mathbf{s}_{v,t}, \hat{\mathbf{s}}_{v,t}),
$
where $\mathbf{s}_{v,t}$ and $\hat{\mathbf{s}}_{v,t}$ denote the ground-truth and predicted scene flow, respectively. 
Compared with $\mathcal{L}_{\mathrm{temp}}$, which supervises coarse point displacement in the DPM space, $\mathcal{L}_{\mathrm{flow}}$ explicitly constrains motion in the Gaussian representation. 
The two losses therefore operate at different levels and are complementary rather than redundant.

\section{Experiments}

\begin{table}[t]
\centering
\caption{\textbf{Comparison to state-of-the-art methods on Waymo (val).} 
PSNR and SSIM are reported. Full: requires dense scene annotations.
Camera: requires camera intrinsics and extrinsics.}
{\footnotesize\setlength{\tabcolsep}{.8mm}
\resizebox{\columnwidth}{!}{
\begin{tabular}{lccccc}
\toprule[1.2pt]                      
\multirow{2}{*}{\textbf{Methods}} & \multirow{2}{*}{\textbf{Supervision}}  & \multicolumn{2}{c}{\textbf{Dynamic-only}}  & \multicolumn{2}{c}{\textbf{Full image}}\\ 
\cmidrule(lr){3-4} \cmidrule(lr){5-6}
& & $\text{PSNR}\uparrow$ & $\text{SSIM}\uparrow$ & $\text{PSNR}\uparrow$ & $\text{SSIM}\uparrow$\\ 
\midrule
\textbf{Per-scene optimization} \\
3DGS \cite{kerbl20233d} & Full &17.13 &0.267  &25.13 &0.741 \\
DeformableGS \cite{yang2024deformable} & Full &17.10 &0.266  &\textbf{25.29} &\textbf{0.761} \\
\midrule
\textbf{Feed-forward model} \\
GS-LRM\cite{zhang2024gs} & Camera &20.02 &0.520 &25.18 &0.753 \\
STORM\cite{yang2024storm} & Camera &\textbf{21.26} &\textbf{0.535} &25.03 &0.750 \\
DynamicVGGT & Image-only &18.07 &0.376 &24.07 &0.676 \\
\bottomrule[1.2pt]
\end{tabular}
}
}
\label{tab:psnr}
\end{table}

This section compares our method to state-of-the-art approaches across multiple tasks to show its effectiveness.

\subsection{Implementation Details}

% By default, we use $L = 12$ layers of MTA, resulting in a total of approximately 1.4B parameters.
% DynamicVGGT is initialized from the pretrained VGGT weights, and about 800M parameters (excluding frozen modules) are fine-tuned in two stages.
% In Stage 1, we train for 10 epochs using the AdamW optimizer with a hybrid schedule—linear warm-up for the first 0.5 epoch followed by cosine decay—peaking at a learning rate of $1\times10^{-6}$.
% In Stage 2, we fine-tune for an additional 50 epochs with the Gaussian head enabled, using the same schedule but a higher peak learning rate of $5\times10^{-5}$.
% All input frames, depth maps, and point maps are resized such that the longer image side does not exceed 518 pixels. The variable temporal step $\Delta t$, randomly sampled within a preset range from 1 to 3. We employ a dynamic batch sizing approach similar to VGGT, processing 18 images
% per GPU. For the training objective, we set $\lambda_1$ = 0.01, $\lambda_2$=0.1, and $\lambda_3$=0.01.
% Training is conducted on 32 $\times$ H200 GPUs, taking about 6 hours for Stage 1 and a day for Stage 2.
% Further details of the training settings and dataset configurations are provided in the Appendix.

% 删减版本
We use $L=12$ MTA layers, resulting in approximately 1.4B parameters. 
DynamicVGGT is initialized from pretrained VGGT weights, with about 800M parameters (excluding frozen modules) fine-tuned in two stages.

In Stage~1, we train for 10 epochs using AdamW with a hybrid learning rate schedule: a linear warm-up for the first 0.5 epoch followed by cosine decay, with a peak learning rate of $1\times10^{-6}$. 
In Stage~2, we further fine-tune the model for 50 epochs with the Gaussian head enabled, using the same schedule but a higher peak learning rate of $5\times10^{-5}$.

All input frames, depth maps, and point maps are resized so that the longer image side does not exceed 518 pixels. 
The temporal offset $\delta$ is randomly sampled from 1 to 3. 
We adopt a dynamic batch sizing strategy similar to VGGT, processing 18 images per batch.
For the training objective, we set 
$\lambda_{\mathrm{temp}} = 0.01$, 
$\lambda_{\mathrm{gs}} = \lambda_{\mathrm{dis}} = 0.1$, 
and $\lambda_{\mathrm{flow}} = 0.01$. Further details of the training settings and dataset configurations are provided in the Appendix.

\subsection{Training Data}
The model is trained on a collection of dynamic autonomous driving datasets, including Waymo Open Dataset~\cite{waymo}, Virtual KITTI~\cite{virtualkitti}, and MVS-Synth~\cite{mvssynth}.
In the first stage, DynamicVGGT is trained on Virtual KITTI and MVS-Synth to learn robust geometric priors and temporal consistency under controlled synthetic settings.
In the second stage, the model is fine-tuned with the Dynamic 3D Gaussian Splatting module on Waymo and Virtual KITTI to enhance dynamic geometry refinement and appearance consistency in real driving environments.
Evaluation is conducted on the Waymo validation set and the KITTI~\cite{kitti17} dataset to assess reconstruction quality.

% More implementation and evaluation details are provided in the Appendix.
\subsection{Point Map Reconstruction}
We evaluate point map reconstruction on the KITTI and Waymo datasets, as shown in Table~\ref{tab:point-estimation}.
We report Accuracy (Acc.), Completion (Comp.), and Normal Consistency (NC). On the KITTI dataset, which uses monocular input with three consecutive frames per sequence, DynamicVGGT achieves the best results across most metrics, obtaining an accuracy of 0.901 and a normal consistency of 0.939.
It consistently outperforms both VGGT~\cite{wang2025vggt} and StreamVGGT~\cite{zhuo2025streaming}, demonstrating its effectiveness in capturing dynamic geometry and maintaining temporal consistency in monocular sequences.

On the Waymo dataset, which provides synchronized multi-view images from three cameras with a frame stride of four, our model generalizes well to large-scale dynamic driving scenes.
It achieves an accuracy of 4.021 and a normal consistency of 0.603.
These results confirm that the proposed dynamic formulation effectively enhances cross-view consistency and scene completeness, even under challenging real-world motion and illumination variations, highlighting the scalability of our feed-forward framework for dynamic 4D perception.

\subsection{4D scene reconstruction}

We further evaluate DynamicVGGT on 4D scene reconstruction using the Waymo validation set, as summarized in Table~\ref{tab:psnr}.
We compare two categories of methods: per-scene optimization and feed-forward models.
DynamicVGGT achieves a PSNR of 18.07 and an SSIM of 0.376 on dynamic regions, and reaches 24.07 and 0.676 on the full-frame evaluation.
Although methods such as STORM obtain higher scores with multi-camera inputs and geometric priors, achieving 21.26 in PSNR and 0.535 in SSIM, DynamicVGGT delivers competitive results using only monocular images without relying on camera parameters or scene-specific optimization.
These results demonstrate that DynamicVGGT effectively reconstructs dynamic 4D scenes with strong temporal consistency and high visual fidelity through purely image-based self-supervision.

\subsection{Monocular and MVS depth estimation}
\begin{table}[t]
\centering
\caption[Valori medi]{Monocular and MVS depth estimation.}
{\footnotesize\setlength{\tabcolsep}{.8mm}
\resizebox{\columnwidth}{!}{
{\begin{tabular}{lcccccc}
\toprule[1.2pt]                      
\multirow{2}{*}{\textbf{Methods}} & \multicolumn{2}{c}{\textbf{{KITTI(Mono)}}}  & \multicolumn{2}{c}{\textbf{{NYU-v2(Mono)}}} & \multicolumn{2}{c}{\textbf{{KITTI(MVS)}}}
\\ 
\cmidrule(lr){2-3} \cmidrule(lr){4-5} \cmidrule(lr){6-7}
          & $ \text{Abs Rel} \downarrow $   & 
          $ \delta<1.25 \uparrow $          & 
          $\text{Abs Rel} \downarrow  $     & 
          $ \delta<1.25 \uparrow $       & 
          $\text{Abs Rel} \downarrow  $     & 
          $ \delta<1.25 \uparrow $            
\\ 
\midrule
DUSt3R\cite{wang2024dust3r}          & 0.109             & 0.873              & 0.081            & 0.909      &0.143 &0.814     \\
MASt3R\cite{leroy2024grounding}          & 0.077             & \textbf{0.948}  & 0.110            & 0.865      &0.115 &0.848     \\
MonST3R \cite{zhang2024monst3r}         & 0.098             & 0.895              & 0.094            & 0.887      &0.107 &0.884      \\
VGGT\cite{wang2025vggt}            & 0.082             & 0.938              & 0.059            & 0.951      &0.062 &0.969     \\
StreamVGGT \cite{zhuo2025streaming}     & 0.082             & 0.947              & \textbf{0.057}            & \textbf{0.959}     &0.173 &0.721       \\
% $\pi^3$         & \textbf{0.060}    & 0.971              & 0.054            & 0.956           \\
DynamicVGGT     & \textbf{0.070} & 0.940  & 0.064            & 0.943     &\textbf{0.051} &\textbf{0.976}      \\

\bottomrule[1.2pt]
\end{tabular}
}
}
}
\label{tab:mono-depth-eval}
\end{table}
\begin{table*}[t]
\centering
\caption[Valori medi]{\textbf{Ablation study.} We evaluate ablated variants of DynamicVGGT on point map estimation over KITTI and Waymo (val). KITTI uses monocular input with three consecutive frames, while Waymo uses 3 frames (stride 4) from the FRONT, SIDE LEFT, and SIDE RIGHT cameras, yielding 9 images per sample. Metrics include Accuracy (Acc.), Completeness (Comp.), Normal Consistency (NC).}
{\footnotesize\setlength{\tabcolsep}{.8mm}
{\begin{tabular}{lcccccccccccccccccc}
\toprule[1.2pt]                      
\multirow{3}{*}{\textbf{Methods}} & \multicolumn{6}{c}{\textbf{{KITTI(Mono)}}}   & \multicolumn{6}{c}{\textbf{{Waymo(3cam)}}}
\\ 
\cmidrule(lr){2-7} \cmidrule(lr){8-13} 
& \multicolumn{3}{c}{\textbf{{Mean}}} & \multicolumn{3}{c}{\textbf{{Med.}}} &
\multicolumn{3}{c}{\textbf{{Mean}}} & \multicolumn{3}{c}{\textbf{{Med.}}}
\\ 
\cmidrule(lr){2-4} \cmidrule(lr){5-7} \cmidrule(lr){8-10} \cmidrule(lr){11-13} 
          & $ \text{Acc.} \downarrow $     & 
          $ \text{Comp.} \downarrow  $     &
          $ \text{NC} \uparrow  $     & 
          $ \text{Acc.} \downarrow   $     & 
          $ \text{Comp.} \downarrow    $     & 
          $ \text{NC} \uparrow   $     & 
          $ \text{Acc.} \downarrow   $     & 
          $ \text{Comp.} \downarrow    $     &
          $ \text{NC} \uparrow  $     & 
          $ \text{Acc.} \downarrow $     & 
          $ \text{Comp.} \downarrow  $     &
          $ \text{NC} \uparrow  $     &   
\\ 
\midrule
Baseline & 1.489 & 0.690 & 0.918 & 1.329 & 0.535 & \textbf{0.971}    & 4.635 & 2.667 & 0.561 & 2.634 & 1.734 & 0.590 \\
 \quad + TA \& FPH(stage1) &0.927 & 0.600 & 0.915 & 0.857  & 0.474 &0.932 & 4.330 & 2.939 & 0.561 & 2.224  & 1.649 & 0.593 \\
 \quad + DGSHead(stage2)  &\textbf{0.901} &\textbf{0.584} &\textbf{0.939} &\textbf{0.733} & \textbf{0.464} &\underline{0.963}   &\textbf{4.021} &2.390 &\textbf{0.562} &\textbf{1.971} &\textbf{1.564} &\underline{0.603} \\
\bottomrule[1.2pt]
\end{tabular}
}
}
\label{tab:ab-study}
\end{table*}

\begin{figure}[t]
\centering
  \includegraphics[width=1\columnwidth]{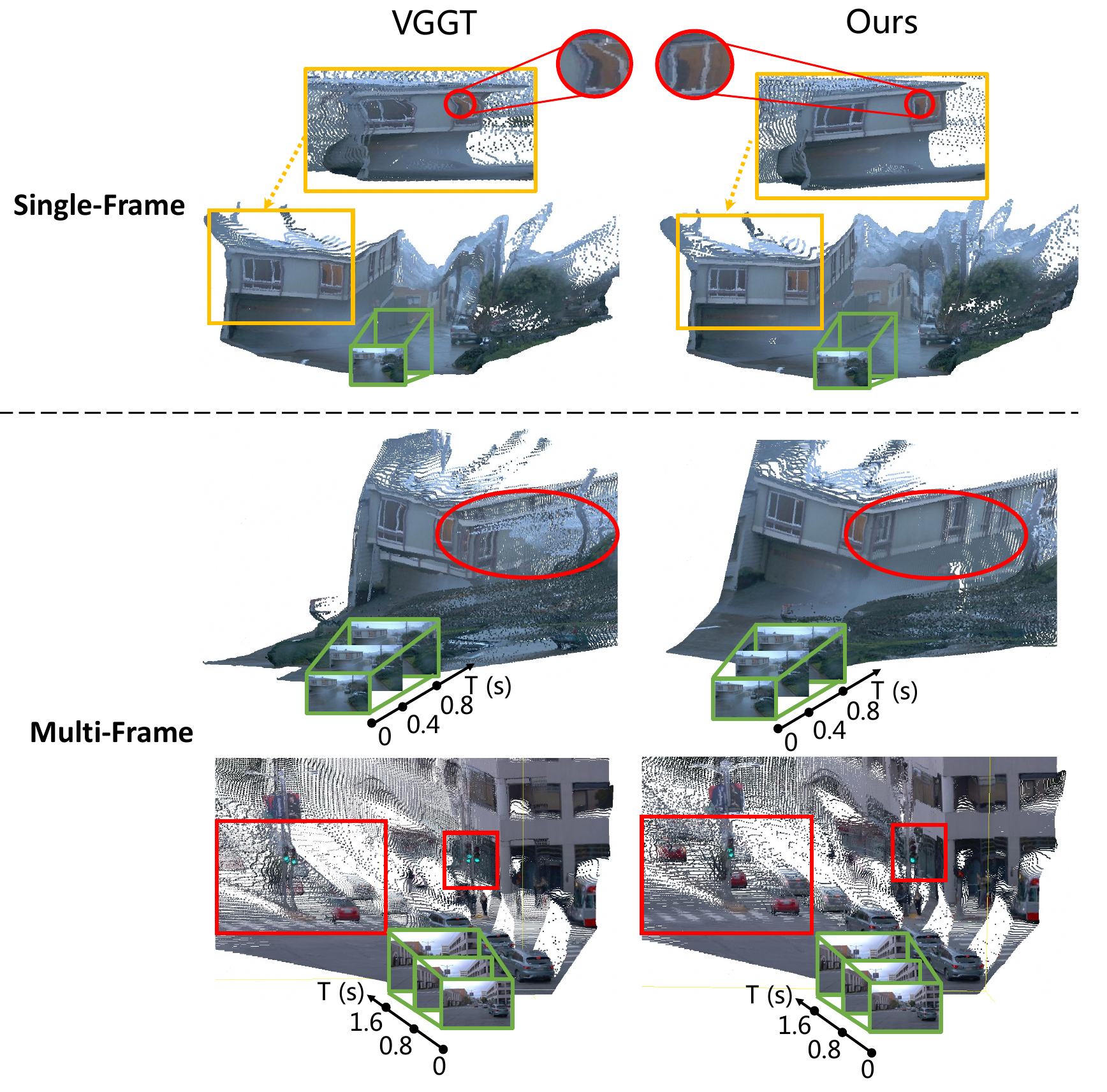}
  \caption{\textbf{Point map reconstruction.} 
DynamicVGGT reconstructs denser, smoother, and more geometrically consistent point maps than VGGT, maintaining temporal coherence even under large viewpoint or scene changes. Zoom in for better view.}
  \label{fig:point_est}
\end{figure}

\begin{figure*}[t]
\centering
  \includegraphics[width=\textwidth]{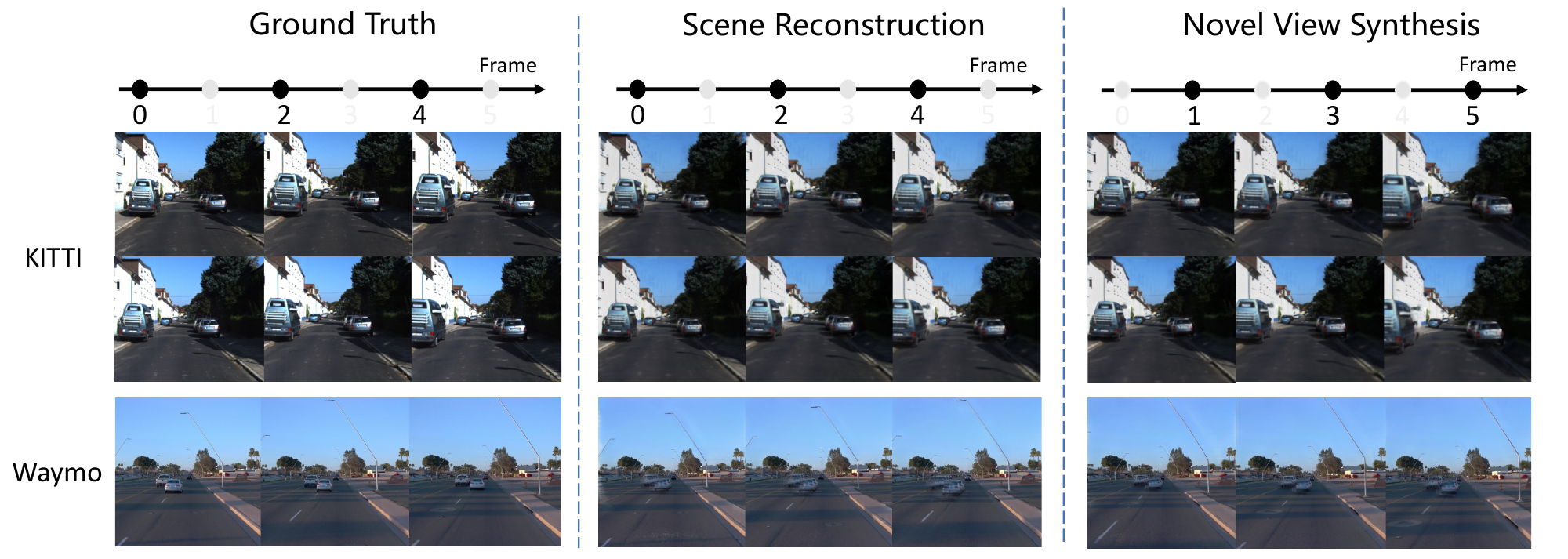}
  \caption{\textbf{Scene Reconstruction and Novel View Synthesis.} Given input frames 0, 2, and 4, DynamicVGGT reconstructs the corresponding scenes and synthesizes novel views for the next frame. For both KITTI and Waymo, multi-view inputs are used; for Waymo, we visualize the front-camera results due to limited view overlap. The model achieves high-quality reconstruction and realistic novel view generation across dynamic driving scenes.}
  \label{fig:3dgs}
\end{figure*}

We evaluate the monocular and multi-view stereo depth estimation performance of DynamicVGGT on three benchmarks: KITTI and NYU-v2~\cite{SilbermanECCV12}, as shown in Table~\ref{tab:mono-depth-eval}. 
We report two standard metrics: Absolute Relative Error (Abs Rel) and accuracy under the $\delta<1.25$ threshold. 

On monocular KITTI, DynamicVGGT achieves an Abs Rel of 0.070, outperforming all baselines. 
On NYU-v2, it obtains an Abs Rel of 0.064 and 94.3\% accuracy under $\delta<1.25$, demonstrating strong generalization from outdoor to indoor scenes. 
Under the multi-view stereo setting, DynamicVGGT achieves the best overall results with an Abs Rel of 0.051 and 97.6\% accuracy, surpassing VGGT and StreamVGGT by a clear margin.

% 放可视化，量化放附录
\subsection{Ablution study}

We conduct an ablation study on the KITTI monocular dataset and the Waymo three-camera dataset to evaluate the contribution of each proposed component, as summarized in Table~\ref{tab:ab-study}.
Starting from the vanilla VGGT baseline, adding temporal attention and the future point prediction head improves accuracy from 1.489 to 0.927 and completeness from 0.690 to 0.600 on KITTI, demonstrating the benefit of temporal modeling in capturing dynamic geometry.

Introducing the Dynamic 3D Gaussian Splatting Head further enhances accuracy to 0.901 and normal consistency to 0.939, producing smoother and more complete reconstructions.
On the Waymo dataset, our full model achieves the best overall results with an accuracy error of 4.021 and a normal consistency of 0.603, confirming that the combination of motion-aware temporal attention and dynamic geometry refinement substantially improves temporal coherence and reconstruction quality.

\subsection{Visualization}

\noindent \textbf{Point Map Reconstruction.}  
We provide qualitative comparisons of point map reconstruction results in Fig.~\ref{fig:point_est}, covering three configurations: single-frame reconstruction, short-term multi-frame reconstruction, and long-range temporal reconstruction.
Across all settings, DynamicVGGT consistently outperforms VGGT in both geometric completeness and temporal consistency.
Even in the single-frame case, our model produces denser and smoother geometry with more accurate structural details, whereas VGGT tends to generate distorted geometry when the viewpoint changes.

When extended to multi-frame and long-sequence inputs, DynamicVGGT demonstrates superior robustness in capturing point-level motion trajectories and maintaining consistent global geometry, particularly in challenging scenarios such as downhill roads and open intersections in autonomous driving.
This highlights the model’s ability to recover fine-grained 3D structures and maintain stable temporal coherence under large scene variations.

\noindent \textbf{Scene Reconstruction and Novel View Synthesis.}  
Figure~\ref{fig:3dgs} presents qualitative results of scene reconstruction and novel view synthesis on the real-world KITTI and Waymo datasets.
Given input frames ${0, 2, 4}$, DynamicVGGT reconstructs the corresponding scenes and synthesizes novel views for the subsequent frame, showcasing its ability to model temporal dynamics from purely image-based inputs.
For both datasets, multi-view images are used as input; for Waymo, we visualize the front-camera results due to the limited overlap between different camera views.

Our method faithfully reconstructs dynamic scenes with moving vehicles and illumination changes, while maintaining consistent global geometry and realistic appearance across time.
Notably, DynamicVGGT achieves temporally coherent novel view synthesis, highlighting its effectiveness as a unified feed-forward framework for dynamic 4D perception in autonomous driving.

\section{Conclusion}

We presented DynamicVGGT, a unified feed-forward framework for dynamic 4D scene reconstruction.
By extending VGGT from static geometry perception to temporal dynamics, our model jointly learns geometric and motion representations through Dynamic Point Maps, Motion-aware Temporal Attention, Future Point Head and a Dynamic 3D Gaussian Head.
This design enables the model to capture temporal dependencies, refine geometry through continuous Gaussian optimization, and maintain feed-forward efficiency. Experimental results demonstrate that DynamicVGGT delivers strong temporal consistency on real-world autonomous driving datasets and simultaneously provides reliable by-products, including camera pose estimation, depth prediction, and novel view synthesis.
We believe this direction will push feed-forward 4D reconstruction closer to a unified paradigm for autonomous driving.

\noindent \textbf{Acknowledge} This work is supported  by NSFC General Program (Grant No. 62576110).

\bibliographystyle{myrefs}
\bibliography{main}



\end{document}